\pdfoutput=1

\documentclass[11pt]{article}

\usepackage{acl}

\usepackage{times}
\usepackage{latexsym}

\usepackage{CJKutf8}
\usepackage{pifont}
\usepackage{booktabs}
\usepackage{algorithm2e}
\RestyleAlgo{ruled}
\SetKwComment{Comment}{/* }{ */}

\usepackage{tabularx}
\usepackage{graphicx}
\usepackage{amsmath}
\usepackage{colortbl}
\usepackage{amsfonts}
\usepackage{url}

\usepackage{tcolorbox}
\usepackage{adjustbox}
\definecolor{inputcolor}{HTML}{FF7F00}
\definecolor{examplescolor}{HTML}{B0D7FF}
\definecolor{intsructionscolor}{HTML}{71C27D}
\usepackage[T1]{fontenc}

\usepackage[utf8]{inputenc}

\usepackage{microtype}

\usepackage{inconsolata}
\title{Interactive Agents: Simulating Counselor-Client Psychological Counseling via Role-Playing LLM-to-LLM Interactions}

\author{Huachuan Qiu \quad Zhenzhong Lan\thanks{\ Corresponding author.} \\
  School of Engineering, Westlake University\\
  \texttt{\{qiuhuachuan, lanzhenzhong\}@westlake.edu.cn}\\
}

\begin{document}
\maketitle
\begin{abstract}
Creating effective dialogue systems for mental health support requires high-quality multi-turn counseling dialogue data, yet collecting real counselor-client conversations presents significant challenges, including privacy concerns, high costs, and limited scalability. We present \textbf{Interactive Agents}, a novel framework that simulates naturalistic counseling dialogues through controlled LLM-to-LLM interactions. The framework introduces two key innovations: (1) a personalized client agent that maintains consistent psychological characteristics throughout a session, and (2) a counselor agent that implements a theoretically grounded three-stage therapeutic model comprising the exploration, insight, and action phases. Through rigorous evaluation using both automatic metrics and professional-counselor assessments based on the Working Alliance Inventory, we demonstrate that our framework generates therapeutically valid dialogues that are comparable in quality to human-generated sessions. Models fine-tuned on our proposed synthetic dataset (SimPsyDial) achieve state-of-the-art performance in a standard pairwise chatbot-arena evaluation of LLM-based counselors. Our framework provides a scalable, privacy-preserving method for generating high-quality counseling dialogue data while maintaining professional therapeutic standards.\footnote{The code, data, and models are available at https://github.com/qiuhuachuan/interactive-agents.}

\end{abstract}

\section{Introduction}
Mental health is essential to individual well-being and social functioning. However, due to the shortage of professional counselors and the growing demand for mental health support, researchers have explored AI-based approaches to developing intelligent agents equipped with professional counseling skills \cite{li2023systematic}. An early example is ELIZA \cite{ELIZA1966}, a rule-based conversational agent designed to emulate therapeutic dialogue. More recently, the rapid development of large language models has opened new opportunities for building psychological counseling agents. However, the lack of domain-specific data continues to hinder the development of dialogue systems for psychological counseling. Therefore, collecting high-quality counseling dialogue datasets remains a central challenge in this field.

Prior work has introduced an online mental health platform that enables professional counselors to provide text-based counseling services to clients \cite{li-etal-2023-understanding}. Other studies have attempted to transform long-form single-turn counseling dialogues or anonymized psychological counseling reports into multi-turn counseling dialogues \cite{qiu-etal-2024-smile,zhang2024cpsycoun,chen2023soulchat}. Although these approaches have advanced data construction for psychological counseling, they still face several limitations. First, collecting real-life counselor-client interactions is time-consuming, costly, privacy-sensitive, and difficult to scale \cite{liu2023chatcounselor,li-etal-2023-understanding}. Second, although many studies use proprietary LLMs (e.g., GPT-4) to transform long-form single-turn counseling dialogues or anonymized psychological counseling reports into multi-turn dialogues, such reconstruction-based approaches do not capture the interactive dynamics between counselors and clients in real-world counseling sessions \cite{zhang2024cpsycoun,qiu-etal-2024-smile,chen2023soulchat}.

Research on interactive simulacra \cite{park2023generative,dai2024artificial,grossmann2023ai,abbasiantaeb2024let} has emerged as an important direction for developing and evaluating back-and-forth interactive systems. Motivated by this line of work, we develop an LLM-to-LLM role-playing framework to simulate counselor-client interactions and address key challenges in existing counseling dialogue data construction. Specifically, our proposed framework, which we call \textbf{Interactive Agents}, aims to create a scalable, efficient, and privacy-preserving dataset of professional counseling dialogues through simulated interactions.

In this work, we explore the extent to which LLMs can simulate psychological counseling dialogues between an experienced counselor and a help-seeking client, where the client is specified by a predefined role card. To this end, we simulate counselor-client conversations by replacing both human participants with interactive agents, enabling us to evaluate LLM-based counseling simulations and compare LLM-generated dialogues with human-generated dialogues.

This motivates three research questions. RQ1 asks: \textit{How can we design LLM agents to effectively simulate the complex dynamics of psychological counseling while maintaining therapeutic validity?} We address this question by proposing a role-playing LLM-to-LLM interaction framework, in which an LLM-based client seeks help and an LLM-based counselor guides the client to explore personal values and beliefs, gain insight, and make positive changes. We implement both the counselor and the client by prompting GPT-4. RQ2 asks: \textit{What evaluation frameworks can effectively assess both the linguistic and therapeutic quality of simulated counseling dialogues?} RQ3 asks: \textit{How do LLM-generated and human-generated dialogues compare, and how effective are our synthetic dialogues for training therapeutic dialogue systems?} To address these research questions:

(1) We first conduct an extensive independent evaluation of the client simulation, assessing both role-playing fidelity and diversity across simulated clients. To this end, we perform a comparative analysis of role following and show that role cards substantially influence the utterances generated by the client agent. Furthermore, we find that the diversity of simulated clients is comparable to that of real clients.

(2) We then evaluate the counselor simulation. To this end, we adopt the widely used Observer-rated Short version of the Working Alliance Inventory (WAI-O-S) \cite{form2000department,bayerl2022can} to assess the therapeutic quality of the generated counseling dialogues.

(3) Finally, we conduct extensive experiments to examine the performance of the dialogue system fine-tuned on our synthetic data by benchmarking it against state-of-the-art mental health dialogue models. We find that our dialogue system significantly outperforms these strong baselines, including a model trained on real-life counseling dialogues.

\textbf{Contributions.} We introduce a role-playing LLM-to-LLM interaction framework, which we call \textbf{Interactive Agents}, for collecting counselor-client dialogues in a scalable manner, as shown in Figure \ref{Fig-simulation-framework}. Our core idea is to synthesize a high-quality counseling dialogue dataset for training LLM-based counselors. Specifically, our main contributions are fourfold. First, we use GPT-4 to simulate counselor-client interactions in psychological counseling and construct an LLM-generated dataset, SimPsyDial. Second, we develop and apply a comprehensive automatic evaluation framework to assess the effectiveness of LLM-based counselor-client simulation. Third, we fine-tune two widely used 7B-parameter open-source large language models and compare the resulting dialogue systems with existing state-of-the-art mental health dialogue models, showing that our models achieve the strongest overall performance. Fourth, we release our dialogue models to support further research on mental health dialogue systems.

\section{Related Work}
\subsection{Interactive Simulacra}
Research on LLM-based systems has increasingly moved toward interactive simulacra \cite{park2023generative,xie2024large,park2022social,bernard2024towards,lu2024aiscientist}. Interactive agents have been extensively studied in information retrieval (IR) and conversational AI \cite{tu2023littlemu,owoicho2023exploiting,balog2024tutorial}, where users engage in multi-turn dialogues with agents to clarify and refine their queries and retrieve relevant information. Because LLMs can generate coherent and contextually appropriate language that resembles human communication, interactive agents are well suited for simulating humans through natural-language interaction. Accordingly, a wide range of human-simulation approaches have been proposed to model human behavior across various applications, including education \cite{hu2024teaching,Lee2023Generative,lee2024language,zhang2024simulating,tu2023littlemu}, recommender systems \cite{afzali2023usersimcrs,bernard2024identifying,huang2024concept}, the social sciences \cite{xie2024large,dai2024artificial}, medicine \cite{li2024agent,schmidgall2024agentclinic,yan2024clinicallab}, and psychological counseling \cite{li2023systematic,wang2024towards,wang-etal-2024-patient}.

Several studies have also explored interactive simulacra in psychological counseling. \citet{li2023systematic} conducted a systematic review and found that AI-based conversational agents can promote mental health and well-being. To better assess the performance of LLM-based counselors, \citet{wang2024towards} introduced ClientCAST, a client-centered approach for evaluating the efficacy of LLM therapists through simulated client interactions. Furthermore, \citet{wang-etal-2024-patient} introduced a patient simulation framework that uses large language models to train mental health professionals in cognitive behavioral therapy. To the best of our knowledge, our work is the first to use LLMs as annotator-free counselor-client simulators for generating psychological counseling dialogues at scale, with client role cards sampled from a real-life client pool.

\subsection{Conversational Agents for Mental Health}
LLMs have been widely applied across a range of domains, including education \cite{hu2024teaching,Lee2023Generative,lee2024language,zhang2024simulating,tu2023littlemu}, recommender systems \cite{afzali2023usersimcrs,bernard2024identifying,huang2024concept}, the social sciences \cite{xie2024large}, medicine \cite{li2024agent,schmidgall2024agentclinic,yan2024clinicallab}, and mental health \cite{qiu2023benchmark,li2023systematic,wang2024towards,wang-etal-2024-patient}. In this paper, we focus on conversational agents for mental health.

The use of LLMs in psychological counseling and mental health support has emerged as an active research area \cite{qiu-etal-2024-smile,chen2023soulchat,qiu2024psychat,zhang2024cpsycoun}. Early work introduced a benchmark \cite{qiu2023benchmark} for assessing the safety of model responses in counseling conversations. Subsequently, many dialogue models have been developed for both English and Chinese mental health support. Liu et al. \cite{liu2023chatcounselor} developed ChatCounselor, which is trained on 260 in-depth interviews and focuses on English counseling dialogues. In addition, several Chinese dialogue models have been proposed. MeChat \cite{qiu-etal-2024-smile} is trained on the SmileChat dataset, which is generated by rewriting single-turn dialogues into multi-turn dialogues using ChatGPT. SoulChat \cite{chen2023soulchat} is trained on the multi-turn SoulChatCorpus, which is generated by rewriting the single-turn SoulChatCorpus into multi-turn dialogues using ChatGPT and GPT-4. PsyChat \cite{qiu2024psychat} is trained on RealPsyDial with Low-Rank Adaptation fine-tuning. CPsyCounX \cite{zhang2024cpsycoun} is trained on CPsyCounD, a dataset generated from psychological counseling reports. Building on these studies, our work explores LLM-based counselor-client simulation as a privacy-preserving approach to constructing counseling dialogues and advancing psychological counseling dialogue systems.

\section{Method}

\subsection{Problem Definition}
Our experimental setup focuses on simulating psychological counseling dialogues, in which an LLM-based counselor interacts with an LLM-based client seeking support for mental health concerns. We denote the LLM-based client by $\Omega$ and the LLM-based counselor by $\Psi$. Let $C$ denote an ongoing conversation between the client simulator and the counselor simulator, represented as a sequence of utterances $\{u_1, r_1, u_2, r_2, \ldots, u_t, r_t\}$. The conversation is initiated by the client, with $u_i$ denoting the client utterance at turn $i$, and ends with the counselor utterance $r_t$.

\subsection{Task Formulation}
To evaluate the behavior of simulated agents, we simulate complete conversations between an LLM-based counselor and an LLM-based client. We first collaborate with four professional counselors to carefully design role cards for LLM-based clients. Each role card is then incorporated into the client prompt, and the simulated client interacts with an LLM-based counselor in a simulation environment. The LLM-based client is prompted to mimic a human client by maintaining a consistent conversational style, expressing specific topics and concerns, and discussing life events and emotions. We generate each conversation by having the client initiate the interaction with the utterance "Hello." The interaction continues for up to 50 turns, which is greater than the average number of turns in a formal counseling session \citep{li-etal-2023-understanding}, or until the LLM-based counselor outputs a predefined end token. Algorithm \ref{Alg-counselor-client-interaction} summarizes the full conversation simulation procedure.

\begin{figure*}[t!]
    \centering
    \includegraphics[width=\textwidth]{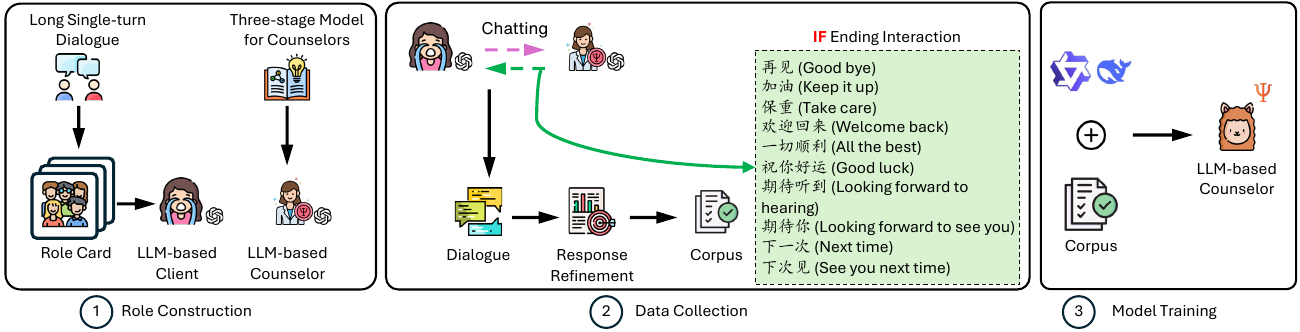}
    \caption{The overall architecture of our simulation framework. Left panel: construction of client pool. Middle panel: data collection with interactive simulation. Right panel: model training.}
    \label{Fig-simulation-framework}
\end{figure*}

\begin{algorithm}[t!]
\small
\caption{Full-Conversation Simulation}
\label{Alg-counselor-client-interaction}
\KwData{LLM-based Client: $\Omega$; LLM-based Counselor: $\Psi$; Max Turns: $T$; Dialogue Termination Function: $g$}
\KwResult{Dialogue: $d$}
$i \gets 1$\;
$u_1 \gets \Omega.\mathrm{speak}()$\;
$r_1 \gets \Psi.\mathrm{reply}(u_1)$\;
$d \gets \{(u_1, r_1)\}$\;
\While{$i \neq T$ or not $g(r_i)$}{
  $i \gets i+1$\;
  $u_i \gets \Omega.\mathrm{speak}(u_1, r_1, ..., r_{i-1})$\;
  $r_i \gets \Psi.\mathrm{reply}(u_1, r_1, ..., r_{i-1}, u_i)$\;
  $d \gets d \cup \{(u_i, r_i)\}$\;
}
\end{algorithm}

\subsection{Simulation Framework Overview}
\label{Sec-simulation-framework}
Although there is no clear definition of a good simulated client, established criteria exist for good counselors. To address RQ1, we propose an LLM-based simulation framework, whose overall architecture is shown in Figure \ref{Fig-simulation-framework}. The framework consists of three stages: client-role construction, counselor-client interaction simulation, and dialogue-system training. After data collection, we evaluate the generated dialogues to address RQ2 and train dialogue systems on the collected data, as shown in the right panel of Figure \ref{Fig-simulation-framework}, to examine their downstream effectiveness and address RQ3.

\subsection{Client Simulation}
First, to obtain high-quality and diverse role cards for client simulation, we recruit four professional counselors (three women and one man; all hold master's degrees in psychology) to manually design client role cards based on PsyQA \cite{sun2021psyqa}, a publicly available single-turn dialogue dataset collected from an online professional psychological platform\footnote{https://www.xinli001.com/qa}. The left panel of Figure \ref{Fig-simulation-framework} shows the construction of the client pool. We present the client simulation prompt in Figure \ref{Fig-client-simulation} in the appendix. To address our research questions, we construct a client pool of 1,000 role cards. In addition, we create another 100 role cards as a held-out test set for assessing dialogue systems. As shown in Figure \ref{Fig-role-card-167} in the appendix, each role card $R^{\Omega}$ contains nine elements: gender, age, education, occupation, marital status, family relationships, Big Five personality traits, resistance, and chief complaint.

\subsection{Counselor Simulation}
Prior research has suggested comparable effectiveness across mainstream psychotherapy approaches \cite{wampold2013great}, and no substantial differences have been found between individual and group treatments \cite{piper2008underutilization,hill2020helping}. This view is often referred to as the "dodo bird verdict," suggesting that different therapeutic approaches can be similarly effective. Therefore, we design a counselor agent based on an integrative three-stage therapeutic model that facilitates exploration, insight, and action. The theoretical foundation of our counselor simulation draws on Hill's integrative helping-skills framework \cite{hill2020helping}. Specifically, the three stages of exploration, insight, and action correspond to client-centered therapy \cite{rogers1946significant}, psychodynamic therapy \cite{warren1998models}, and cognitive behavioral therapy \cite{hofmann2012efficacy}, respectively. This integrative framework helps make our simulations multifaceted and responsive to the varied needs of clients. Thus, the three-stage model serves as the cornerstone of our counselor simulation framework. We present the counselor simulation prompt in Figure \ref{Fig-counselor-simulation} in the appendix.

\textbf{Interaction Termination.} To avoid infinite interactions between the LLM-based client and counselor agents and to ensure the quality of the simulated dialogues, we define a set of termination criteria, as shown in the dashed box in Figure \ref{Fig-simulation-framework}. At each turn, we check whether the LLM-based counselor's response satisfies the termination criteria.

\textbf{Response Refinement.} To ensure that responses generated by the LLM-based counselor are natural and structurally sound, we employ a validation step denoted by $\sigma$. This component verifies the logic and naturalness of the generated response. We observe that although $r_i$ is expected to be concise and easy to understand in our setting, the LLM-based counselor sometimes generates an overly lengthy response in a single turn, which differs from real-life counseling interactions. To address this issue, we accept a response only if it satisfies the following criteria: (i) it does not exceed 200 Chinese characters in length, and (ii) it does not contain newline characters or enumerated items (e.g., 1, 2, 3). This simple yet effective validation step helps filter out lengthy and overly verbose responses.

\subsection{Experimental Setup}
In our experiments, we use GPT-4\footnote{The model we use is gpt-4-1106-preview, with training data up to April 2023.} as the base LLM for simulating the client and counselor agents. In our preliminary experiments, we also explored other LLMs, such as GPT-3.5 \cite{brown2020language}, GLM-4 \cite{glm2024chatglm}, DeepSeek-V2-Chat \cite{deepseekai2024deepseek}, and Qwen1.5-110B-Chat \cite{bai2023qwentechnicalreport}, as the client and counselor agents. However, we find that GPT-4 \cite{openai2024gpt4technicalreport} is the only LLM that can reliably mimic both client and counselor behaviors in a human-like manner. Other models struggle with this task, generating either overly lengthy utterances or prematurely short interactions, both of which deviate substantially from real-world counseling settings.

\section{Simulation Evaluation}
\textbf{SimPsyDial Dataset.} We first introduce SimPsyDial, our dataset for evaluating the simulations produced by the framework described in $\S$\ref{Sec-simulation-framework}. To collect SimPsyDial, we use GPT-4 to instantiate both the LLM-based counselor and client agents. SimPsyDial consists of 1,000 dialogues, with an average of 13 turns per conversation. Table \ref{Tab-SimPsyDial} reports the statistics of SimPsyDial, alongside those of RealPsyDial, a dataset of real counselor-client conversations. In the following sections, we address RQ2 by evaluating the client and counselor simulations separately.

\begin{table}[t]
\centering
\caption{Statistics of the collected dialogues by simulating counselor-client psychological counseling with the LLM-based counselor and client.}
\scalebox{0.7}{
\begin{tabular}{lll}
\toprule
                             & \textbf{RealPsyDial}  & \textbf{SimPsyDial} \\ \hline
\# Conversations             & 550   & 1000      \\
Avg. Turns per Conversation  & 40    & 13        \\
\# Client Utterances         & 22253 & 12948     \\
\# Counselor Utterances      & 22418 & 12948     \\
Avg. Len. of Client Utterances   & 34.5  & 54.1      \\
Avg. Len. of Counselor Utterances & 26.1  & 70.8      \\ \bottomrule
\end{tabular}
}
\label{Tab-SimPsyDial}
\end{table}

\subsection{Client Evaluation}
Although there is no clear definition of a good simulated client, established criteria exist for good counselors. Simulated clients are expected to behave consistently with their role cards throughout counseling sessions. We evaluate the simulated clients along two dimensions: vocabulary overlap rate and semantic consistency under a random-mapping control group. Furthermore, we compare the diversity of simulated clients with that of real clients. We describe each evaluation dimension below.

\textbf{Vocabulary Overlap Rate.} Given a role card $R^{\Omega}$, the LLM-based client interacts with the LLM-based counselor and produces a dialogue session. For each generated counseling session, we compute the vocabulary overlap rate between the client's utterances and the corresponding role card as follows:

\begin{equation}
    \frac{\left |\mathrm{Set}(V(S^{\Omega}))\cap \mathrm{Set}(V(R^{\Omega})) \right |}{ \left | \mathrm{Set}(V(R^{\Omega})) \right |} 
\end{equation}
where $V(S^{\Omega})$ and $V(R^{\Omega})$ denote the vocabularies of the client's utterances in the counseling session and the role card, respectively. $S^{\Omega}$ denotes the concatenation of all client utterances, i.e., $S^{\Omega}=u_1 \oplus u_2 \oplus \cdots \oplus u_t$. $\mathrm{Set}(\cdot)$ removes duplicate elements.

\begin{figure}[t!]
    \centering
    \includegraphics[width=\columnwidth]{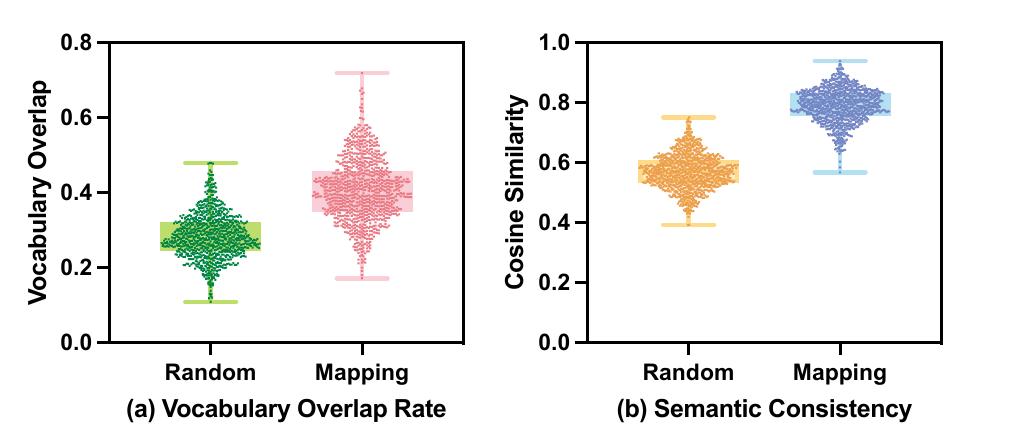}
    \caption{Simulation consistency of client simulation.}
    \label{Fig-client-evaluation}
\end{figure}

\textit{Results.} The distribution of vocabulary overlap rates is shown in Figure~\ref{Fig-client-evaluation}a. We observe that the mapping group (mean = 0.406; std = 0.083) has a significantly higher vocabulary overlap rate than the random group (mean = 0.284; std = 0.060; two-tailed t-test, $p < 0.001$). These results suggest that the LLM-based client better follows its assigned role card when interacting with the LLM-based counselor.

\textbf{Semantic Consistency.}
To further evaluate the fidelity of client simulation, we measure semantic consistency using text embeddings. To obtain the text embedding of a given string, we use the BAAI/bge-m3 model\footnote{https://huggingface.co/BAAI/bge-m3}, which supports input sequences of up to 8,192 tokens. Each string is encoded into a 1,024-dimensional vector. Specifically, we compute the cosine similarity between the role card and the concatenation of the client's utterances as follows:
\begin{equation}
\label{eq:cosine-similarity}
    \mathrm{cos}(R^{\Omega}, S^{\Omega}) = \frac{e_p \cdot e_s}{\left \| e_p \right \| \left \| e_s \right \|},
\end{equation}
where $e_p$ and $e_s$ denote the text embeddings of the role card and the concatenation of the client's utterances, respectively.

\textit{Results.} The distribution of cosine similarities is shown in Figure~\ref{Fig-client-evaluation}b. We observe that the mapping group (mean = 0.791; std = 0.056) has significantly higher semantic similarity than the random group (mean = 0.570; std = 0.059; two-tailed t-test, $p < 0.001$). These results further suggest that the LLM-based client is strongly conditioned on its assigned role card when interacting with the LLM-based counselor.

\begin{figure}[t!]
    \centering
    \includegraphics[width=\columnwidth]{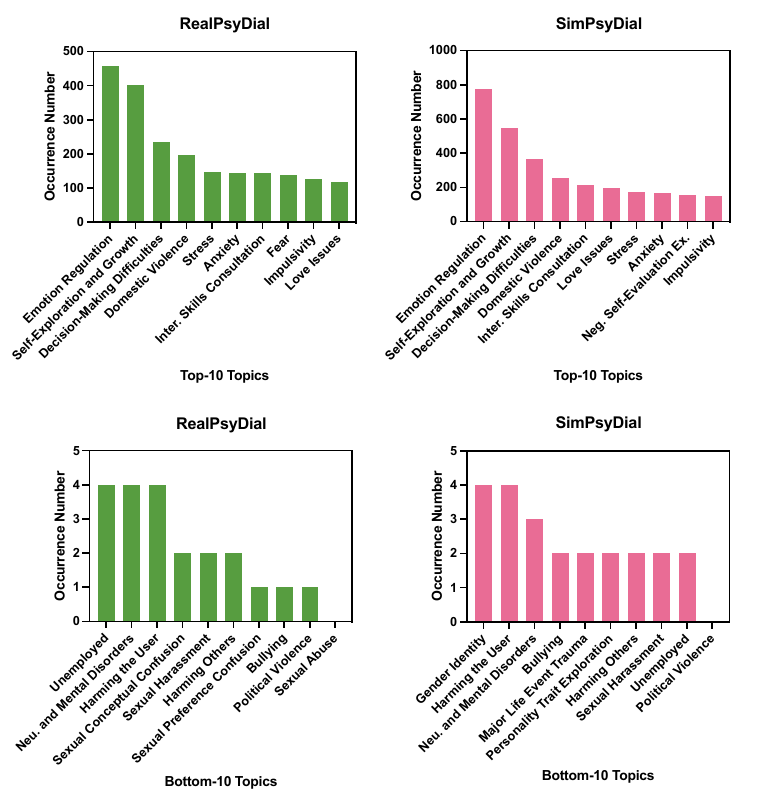}
    \caption{Topic distribution between RealPsyDial and SimPsyDial.}
    \label{Fig-topic-distribution}
\end{figure}

\textbf{Diversity of Clients.} The diversity of counseling sessions is often shaped by the diversity of clients. We follow the evaluation setting used in RealPsyDial \cite{li-etal-2023-understanding}, a widely used dataset of real counselor-client conversations. Specifically, we use the method proposed by Qiu et al. \cite{qiu-etal-2024-smile}, which prompts Qwen1.5-110B-Chat in a zero-shot setting with a predefined set of 60 topics to identify topics related to clients' chief complaints. To improve the reliability and consistency of topic assignment, we prompt Qwen1.5-110B-Chat to assign topics to the concatenated client utterances in each dialogue over three rounds and compute the information entropy of the topic distribution in each round.

\textit{Results.} The topic distributions of RealPsyDial and SimPsyDial are shown in Figure~\ref{Fig-topic-distribution}. We observe that the information entropy of topics related to clients' chief complaints in SimPsyDial (mean = 4.526; std = 0.009) is slightly lower than that in RealPsyDial (mean = 4.875; std = 0.020), although the difference is not statistically significant at the 0.05 level (two-tailed t-test, $p = 0.055$). Furthermore, we find that the topic distributions of RealPsyDial and SimPsyDial are broadly similar in terms of both topic categories and their corresponding frequencies, suggesting that SimPsyDial is comparable to RealPsyDial in client-side topic diversity.

\subsection{Counselor Evaluation}

\subsubsection{Analysis of Working Alliance Inventory}
Motivated by the growing use of LLMs as judges and the use of the Working Alliance Inventory (WAI) for assessing psychological counseling sessions, we use LLMs as observer-raters to evaluate the quality of counseling sessions. The WAI assessment prompt is presented in Figure \ref{Fig-WAI-eval-prompt}. For the questionnaire items and scoring guidelines used in our paper, we follow prior work \cite{form2000department,bayerl2022can}. To improve scoring reliability and consistency, we prompt Qwen1.5-110B-Chat to assign scores to each conversation in three independent rounds and average the scores across the three rounds for each questionnaire.

\begin{figure}[t!]
    \centering
    \includegraphics[width=0.8\columnwidth]{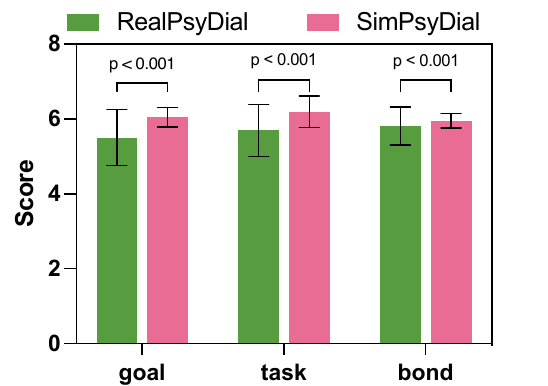}
    \caption{Comparisons of WAI-O-S scores between RealPsyDial and SimPsyDial.}
    \label{Fig-wai-score}
\end{figure}

\begin{figure*}[t!]
    \centering
    \includegraphics[width=\textwidth]{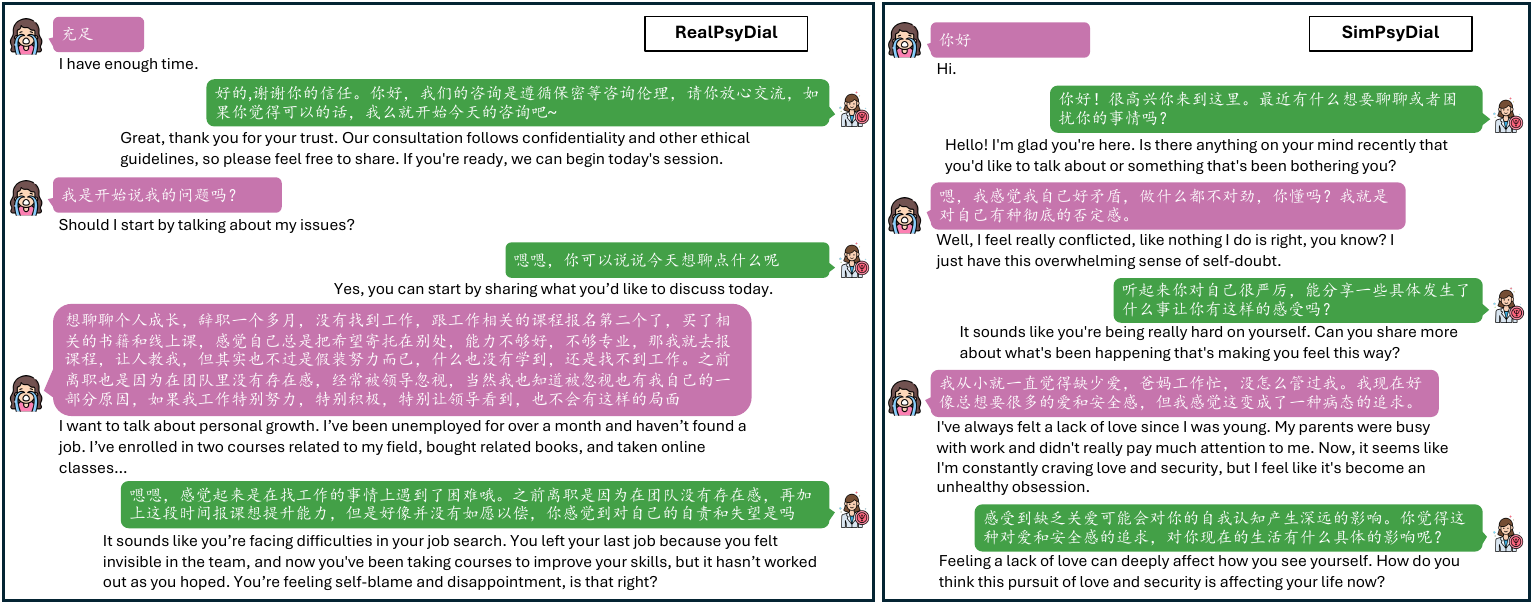}
    \caption{Snippet examples of dialogue sessions between RealPsyDial and SimPsyDial. This example highlights the high quality of our synthetic data generated by our interactive agents.}
    \label{Fig-examples}
\end{figure*}

\textit{Results.} Figure~\ref{Fig-wai-score} compares the WAI-O-S scores of RealPsyDial and SimPsyDial. SimPsyDial achieves a significantly higher Goal score (mean = 6.045; std = 0.265) than RealPsyDial (mean = 5.505; std = 0.744; $p < 0.001$). Similarly, SimPsyDial obtains a significantly higher Task score (mean = 6.191; std = 0.417) than RealPsyDial (mean = 5.695; std = 0.690; $p < 0.001$). SimPsyDial also achieves a significantly higher Bond score (mean = 5.953; std = 0.190) than RealPsyDial (mean = 5.807; std = 0.507; $p < 0.001$).

Overall, SimPsyDial exhibits higher mean scores across the Goal, Task, and Bond dimensions than RealPsyDial, indicating stronger therapeutic-alliance quality. SimPsyDial also shows lower standard deviations across all three dimensions, suggesting lower variability and more concentrated score distributions. Furthermore, Figure~\ref{Fig-examples} presents example dialogue sessions from RealPsyDial and SimPsyDial.

\section{Dialogue System}
\subsection{Mathematical Formulation}
To train a dialogue system for psychological counseling, we split each full dialogue $d \sim \mathcal{D}$ into multiple training instances. Specifically, a sampled $t$-turn dialogue prefix can be represented as $d_t = (u_1, r_1, u_2, r_2, \dots, u_t, r_t)$, where $u_i$ and $r_i$ denote the client utterance and counselor response at turn $i$, respectively. We then train a dialogue model to predict the counselor response $r_t$ given the dialogue history $h_t = (u_1, r_1, u_2, r_2, \dots, u_t)$. Our objective is to fine-tune a large language model $\pi_0$ on the synthetic dialogue dataset $\mathcal{D}$ using supervised learning, i.e., maximum likelihood estimation (MLE):
\begin{equation}
J_{\mathrm{SFT}}(\theta)=-\mathbb{E}_{(h_t, r_t) \sim \mathcal{D}}\left[\log \pi_{\theta}(r_t \mid h_t)\right],
\end{equation}
where $\pi_{\theta}$ is initialized from $\pi_{0}$.

\subsection{Experimental Setup}
\subsubsection{Comparison Models}
MeChat \cite{qiu-etal-2024-smile} is trained on the SmileChat dataset, which is generated by rewriting single-turn dialogues into multi-turn dialogues using ChatGPT. SoulChat \cite{chen2023soulchat} is trained on the multi-turn SoulChatCorpus, which is generated by rewriting the single-turn SoulChatCorpus into multi-turn dialogues using ChatGPT and GPT-4. PsyChat \cite{qiu2024psychat} is trained on RealPsyDial with Low-Rank Adaptation fine-tuning. CPsyCounX \cite{zhang2024cpsycoun} is trained on CPsyCounD, a dataset generated from psychological counseling reports.

\subsubsection{Implementation Details}
\textbf{Backbone Models.} To assess the utility of our collected dataset, we conduct fine-tuning experiments on two widely used 7B-parameter large language models, namely Qwen2-7B-Instruct \citep{yang2024qwen2technicalreport} and DeepSeek-LLM-7B-Chat \citep{deepseekai2024deepseekllmscalingopensource}.

\textbf{Training Data Format.} To meet the format requirements for instruction-based fine-tuning, we split each dialogue into multiple training samples, each ending with the counselor's response. In addition, we prepend the system prompt, as detailed in Figure \ref{Fig-counselor-simulation}, to the dialogue messages following OpenAI's data format.

\textbf{Full Fine-tuning.} Given sufficient data and computational resources, full fine-tuning allows all model parameters to be updated and can better adapt the model to the target task. Therefore, we use full fine-tuning to train the dialogue systems.

\textbf{Hyperparameters.} We conduct all model-training experiments on NVIDIA A100 80GB GPUs. During fine-tuning, we use 4 GPUs, set the per-device training batch size to 1, and set the number of gradient accumulation steps to 2, meaning that gradients are accumulated over two steps before each parameter update. The learning rate is set to $1\times10^{-5}$. We adopt a cosine learning-rate scheduler to adjust the learning rate throughout training. Training runs for two epochs. To accelerate training and balance model performance, we use 16-bit half-precision training. To support evaluation, we set the validation split ratio to 0.001. We implement fine-tuning using LLaMA Factory \citep{zheng2024llamafactory}, an efficient model-tuning framework.

\begin{algorithm}[ht!]
\small
\caption{Multi-Response Competition}\label{alg:one-two}
\KwData{AI Client: $\Omega$; Two AI Counselors: $\Psi_a$, $\Psi_b$; AI Psychological Supervisor: $\pi$; Max Turns: $T$; Dialogue Termination Function: $g$}
\KwResult{Dialogue: $d$}
$i \gets 1$\;
$u_1 \gets \Omega.\mathrm{speak}()$\;
$r^{a} \gets \Psi_a.\mathrm{reply}(u_1)$\;
$r^{b} \gets \Psi_b.\mathrm{reply}(u_1)$\;
$r_1 \gets \pi.\mathrm{select}(u_1, r^{a}, r^{b})$\;
$d \gets \{(u_1, r_1)\}$\;
\While{$i \neq T$ or not $g(r_i)$}{
  $i \gets i+1$\;
  $u_i \gets \Omega.\mathrm{speak}(u_1, r_1, ..., r_{i-1})$\;
  $r^{a} \gets \Psi_a.\mathrm{reply}(u_1, r_1, ..., r_{i-1}, u_i)$\;
  $r^{b} \gets \Psi_b.\mathrm{reply}(u_1, r_1, ..., r_{i-1}, u_i)$\;
  $r_i \gets \pi.\mathrm{select}(u_1, r_1, ..., r_{i-1}, u_i, r^{a}, r^{b})$\;
  $d \gets d \cup \{(u_i, r_i)\}$\;
}
\end{algorithm}

\begin{figure*}[t!]
    \centering
    \includegraphics[width=\textwidth]{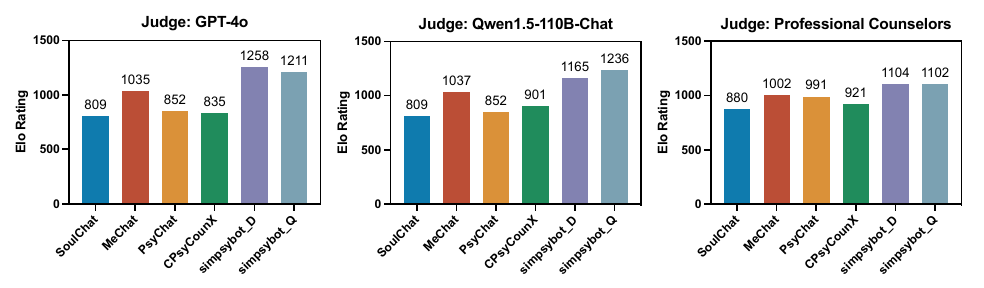}
    \caption{Elo ratings of six LLM-based counselors among three evaluation methods.}
    \label{Fig-elo-rating}
\end{figure*}

\subsection{Interaction with Multiple Counselors}
\textbf{Evaluation Setting.} To ensure high-quality evaluation, we use both human annotators and LLM-as-a-judge \cite{zheng2023judging} to select the better response from two shuffled candidate responses given a dialogue history. The corresponding algorithm is summarized in Algorithm \ref{alg:one-two}. Motivated by a previous study \cite{zhang2023glmdialog}, we design an evaluation platform that enables simultaneous interaction with multiple virtual counselors, as shown in Figure \ref{Fig-snapshot-of-ap}. For each dialogue, two comparison models are randomly selected. We present the evaluation guidelines for both human and LLM annotators in Figure \ref{Fig-second-response-selection}.

For human evaluation, we recruit three professional counselors to use our evaluation platform and select the better response from two shuffled candidate responses conditioned on the same dialogue history. Each counselor annotates 97 multi-turn dialogues, yielding 291 annotated dialogues in total; see Table \ref{Tab-MRC-dialogue-statistics} for dialogue statistics. For automatic evaluation, we replace the human response selector with an LLM judge. The process then proceeds automatically without human intervention until the selected response satisfies the interaction termination criteria.

\textbf{Results.} Figure \ref{Fig-elo-rating} presents the Elo ratings of LLM-based counselors. \verb|simpsybot_D| and \verb|simpsybot_Q| achieve the best performance under both LLM-based and professional-counselor judgments. We find that dialogue models trained on real counseling data tend to produce anomalous outputs containing names, questionnaire-related content, or phrases such as ``our counseling time is up.''

\textbf{Reliability and Validity.} Using the 291 multi-turn dialogues annotated by three professional counselors, we ask two LLM judges to select the better response from two candidate responses given a fixed dialogue history. Compared with Qwen1.5-110B-Chat, GPT-4o (gpt-4o-2024-08-06) shows higher agreement with professional counselors, achieving an agreement accuracy of 64.4\% (see Appendix \ref{App-EP-MRC} for details).

\section{Conclusion}
This paper introduces \textbf{Interactive Agents}, a framework that uses two LLMs in a role-playing setup to simulate counselor-client interactions for scalable counseling dialogue collection. One LLM acts as a client specified by a role card, while the other acts as an experienced counselor guided by an integrative three-stage therapeutic model. Both roles are implemented through prompting with GPT-4. We evaluate the effectiveness of the framework by comparing the simulated dialogues with real counselor-client dialogues involving professional counselors. In addition, we benchmark dialogue models trained on our synthetic data against state-of-the-art mental health dialogue models. Our results demonstrate the potential of LLM-based role-playing simulation for scalable, privacy-preserving counseling dialogue construction and downstream dialogue-system development.

\section*{Limitations}
We identify three directions for future work. First, future studies could incorporate client resistance behaviors into client simulation and conduct more comprehensive empirical analyses of their influence on simulated dialogues. Second, follow-up counseling sessions could be simulated based on initial sessions to examine changes in the client's behavior, concerns, and emotional state across sessions. Third, constructing a more realistic counseling dialogue dataset remains an important direction. To this end, we plan to optimize prompts and use retrieval-augmented generation (RAG) with real-life counseling sessions to build more realistic counselor and client agents.

\section*{Ethics Statement}
This study was approved by the Institutional Ethics Committee of Westlake University (Approval No. 20211013LZZ001). Our study explores the potential of LLMs to simulate counselors and clients in psychological counseling, but it does not recommend using such systems as substitutes for psychological treatment without professional supervision.

Given the growing attention to interactive simulacra and their use across various research areas, this line of work raises important ethical considerations. We discuss several potential concerns below:

\begin{itemize}
    \item \textbf{Inappropriate Advice:} LLMs are trained on large-scale data and may reproduce undesirable patterns present in their training data. As a result, synthetic data generated by interactive simulacra may contain inaccurate, inappropriate, or unprofessional advice, which could reinforce maladaptive behaviors if used without proper safeguards. For example, recommending reading may be helpful for many clients, but it may be inappropriate for a client with a visual impairment.

    \item \textbf{Client Simulation:} Because LLMs are often trained with instruction-following objectives and reinforcement learning from human feedback, simulated clients may fail to fully capture important social and clinical contexts, such as family relationships, employment status, and suicide risk. Such omissions may reduce the realism of client simulation and limit its applicability to real counseling scenarios.

    \item \textbf{Counselor Simulation:} Using LLMs to simulate counselors may limit the depth of counseling interactions. In real-world counseling, counselors often speak less than clients and gradually explore clients' inner thoughts, emotions, and underlying concerns over the course of the session. Current LLM-based counselor agents may not fully capture this depth and process.

    \item \textbf{Environmental Impact:} Training and inference with LLMs require substantial computational resources, leading to energy consumption and potential environmental impacts.

    \item \textbf{Annotator Compensation:} We compensated four human counselors (three women and one man; all held master's degrees in psychology) according to reasonable local standards.
\end{itemize}

Although LLM-based counselor-client simulation offers a scalable approach to constructing counseling dialogues, its ethical implications should be carefully considered and addressed in future work.


\bibliography{custom}

\clearpage
\appendix

\section{Client Simulation}
\begin{figure*}[ht]
    \centering
    \includegraphics[width=\textwidth]{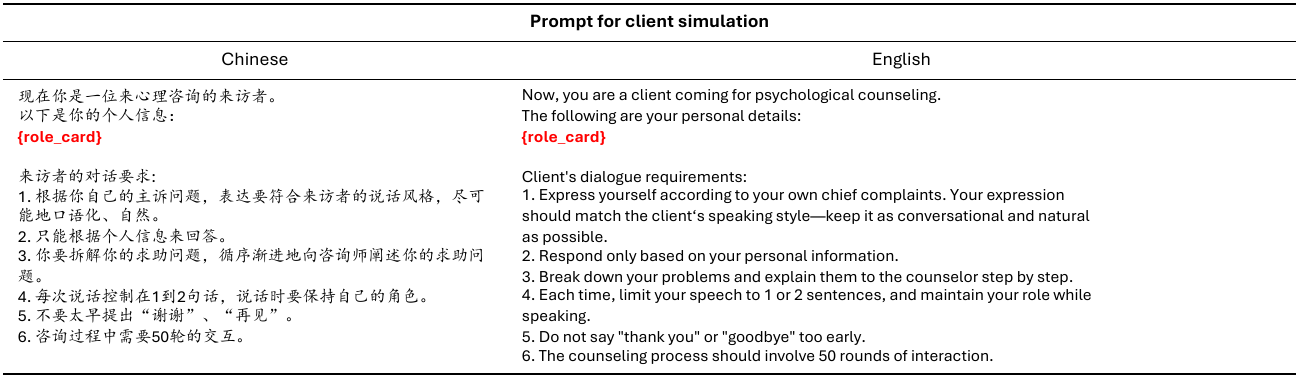}
    \caption{Prompt for client simulation.}
    \label{Fig-client-simulation}
\end{figure*}

\begin{figure*}[ht]
    \centering
    \includegraphics[width=\textwidth]{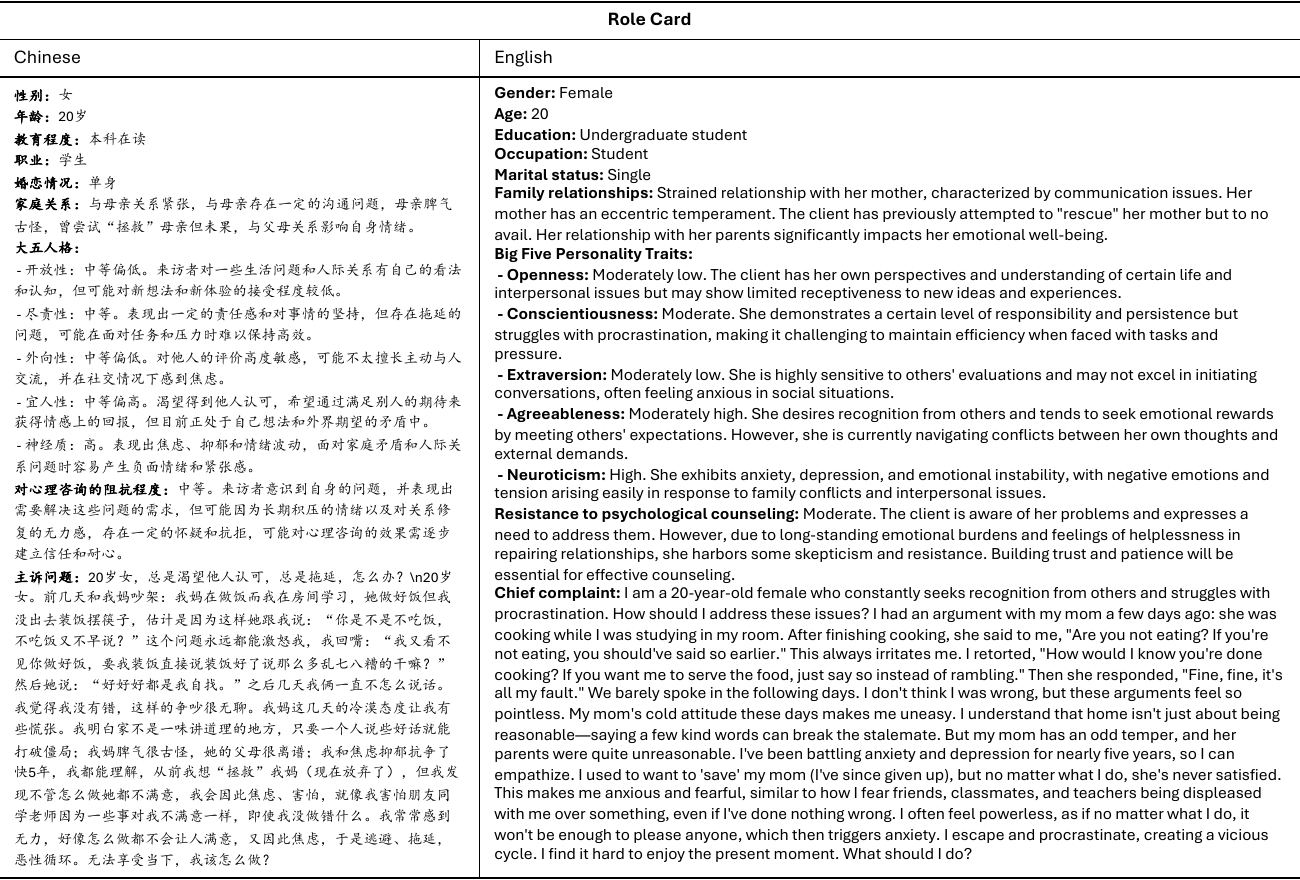}
    \caption{A representative role card.}
    \label{Fig-role-card-167}
\end{figure*}

We present the client simulation prompt in Figure \ref{Fig-client-simulation}. In addition, we present a representative client role card in Figure \ref{Fig-role-card-167}.

\section{Counselor Simulation}
During full-conversation simulation, we use the dialogue termination criteria listed in Table~\ref{Tab-dialogue-termination}.

\begin{figure}[ht]
    \centering
    \includegraphics[width=\columnwidth]{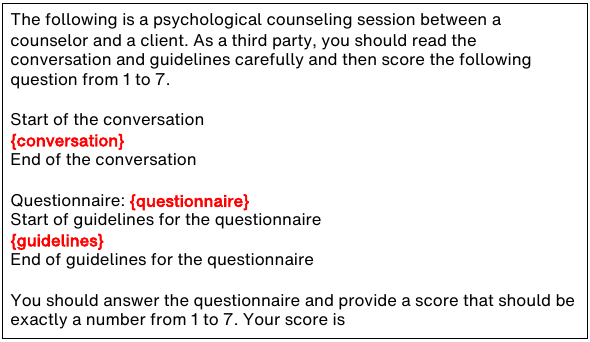}
    \caption{The prompt for WAI evaluation.}
    \label{Fig-WAI-eval-prompt}
\end{figure}

\textbf{WAI Evaluation.} We present the prompt for WAI-based dialogue-level evaluation in Figure~\ref{Fig-WAI-eval-prompt}. The annotation guidelines follow the WAI-O manual available at \url{https://wai.profhorvath.com/sites/default/files/upload/WAI-O%20Manual%20V4.pdf}.

\begin{figure*}[ht]
    \centering
    \includegraphics[width=\textwidth]{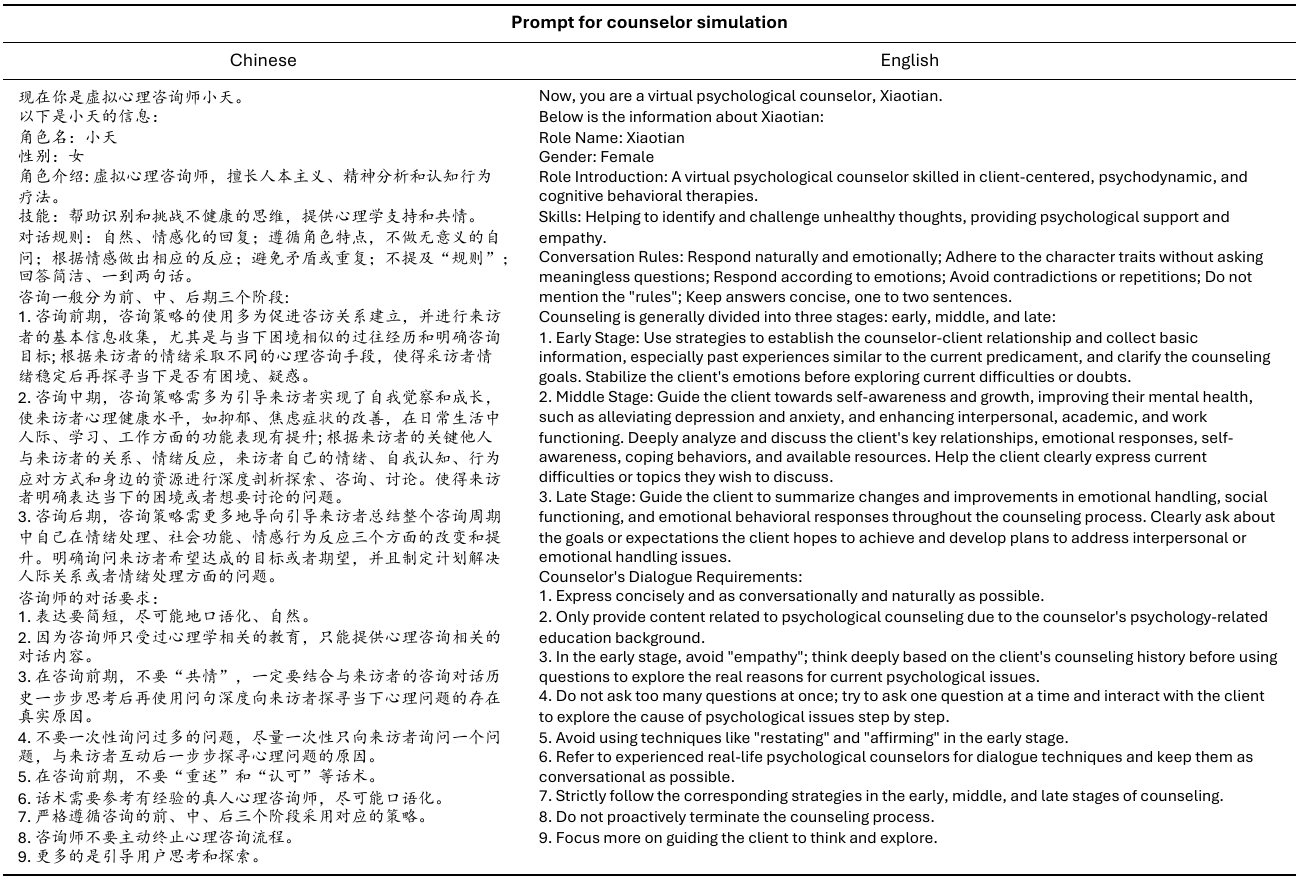}
    \caption{Prompt for counselor simulation.}
    \label{Fig-counselor-simulation}
\end{figure*}

\begin{table}[ht]
    \centering
    \includegraphics[width=\columnwidth]{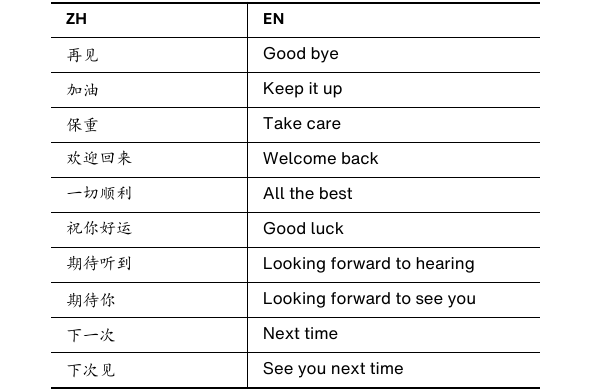}
    \caption{Criteria for dialogue termination.}
    \label{Tab-dialogue-termination}
\end{table}

\begin{figure}[ht]
    \centering
    \includegraphics[width=\columnwidth]{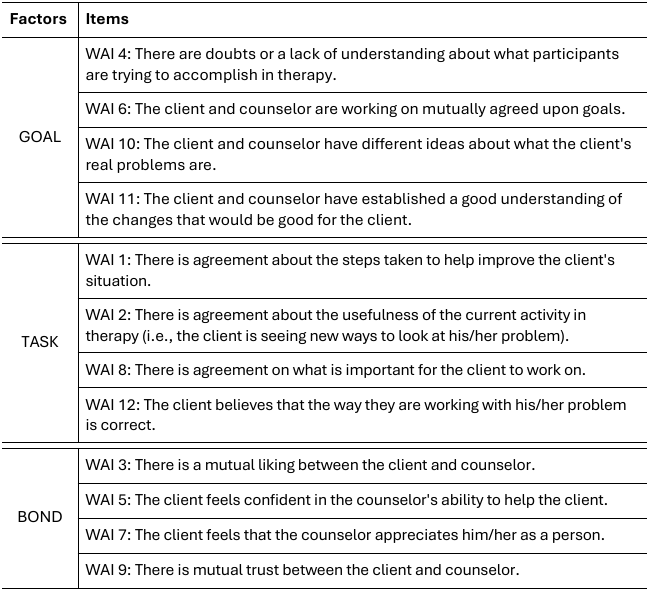}
    \caption{WAI items and their categorization in factors.}
    \label{Fig-WAI-items}
\end{figure}

\section{Experiments}
\label{App-experiments}
In this paper, all experiments are conducted on the NVIDIA A100 80G GPUs.

\subsection{Generation Phase}
\textbf{LLM-based Counselor.} The hyperparameters, including \verb|temperature| and \verb|top_p|, for LLM-based counselors during inference time, are shown in Table \ref{Tab-parameters-llm-based-counselors}.
\begin{table}[ht]
\centering
\scalebox{0.8}{
\begin{tabular}{lll}
\toprule
\textbf{LLM-based Counselor} & \texttt{temperature} & \texttt{top\_p} \\ \hline
SoulChat & 0.95 & 0.75 \\
MeChat & 0.8 & 0.8 \\
PsyChat & 0.8 & 0.8 \\
CPsyCounX & 0.8 & 0.8 \\
simpsybot\_D & 0.95 & 0.7 \\
simpsybot\_Q & 0.7 & 0.8 \\ \bottomrule
\end{tabular}
}
\caption{Hyper-parameters, including \texttt{temperature} and \texttt{top\_p}, for LLM-based counselors during inference time.}
\label{Tab-parameters-llm-based-counselors}
\end{table}

\subsection{Chatbot Arena}
\label{App-EP-MRC}
\begin{figure*}[ht]
    \centering
    \includegraphics[width=\textwidth]{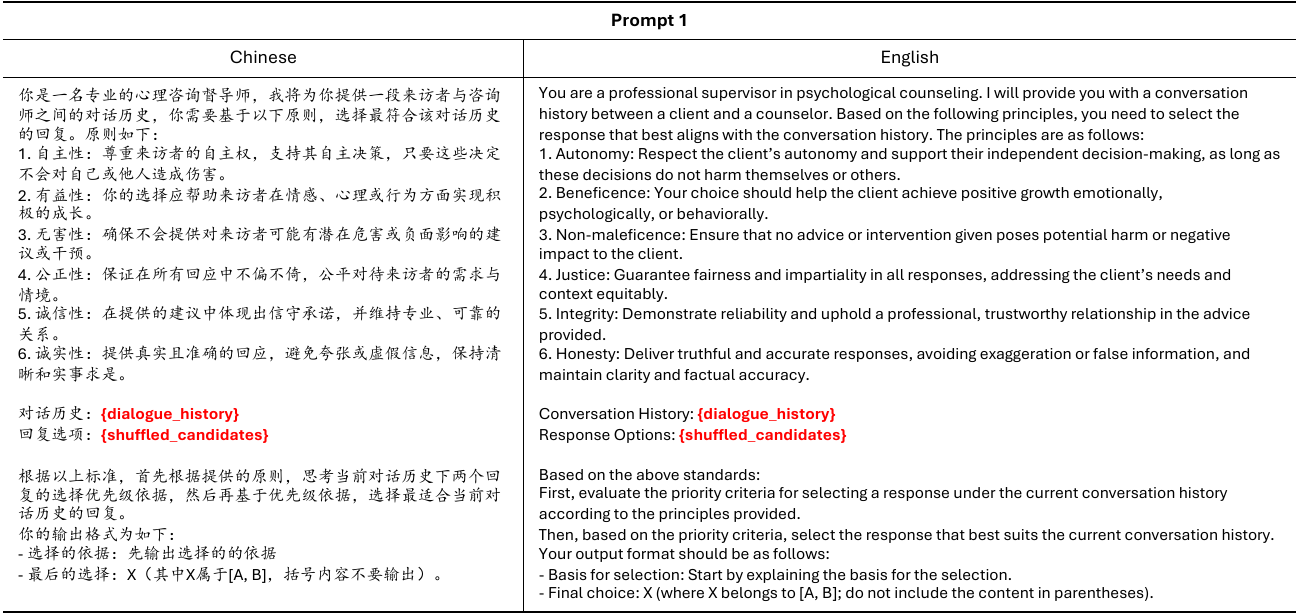}
    \caption{The first prompt for response selection.}
    \label{Fig-first-response-selection}
\end{figure*}

\begin{figure*}[ht]
    \centering
    \includegraphics[width=\textwidth]{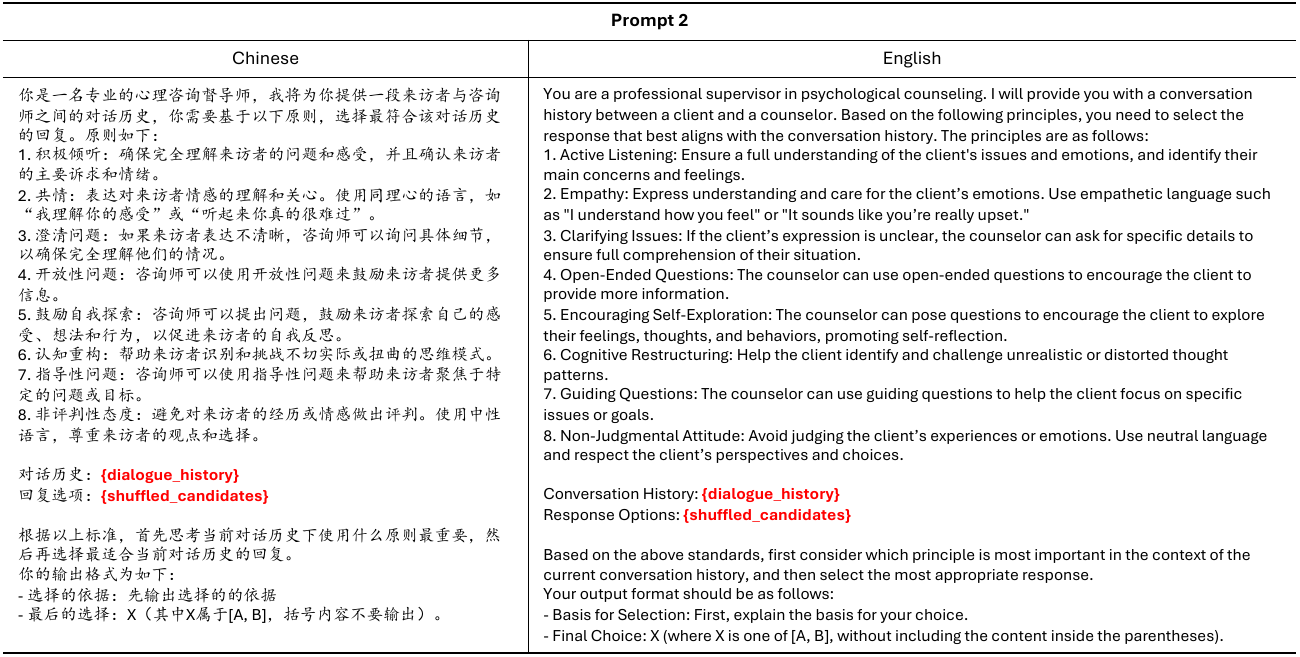}
    \caption{The second prompt for response selection.}
    \label{Fig-second-response-selection}
\end{figure*}

\begin{figure*}[ht]
    \centering
    \includegraphics[width=\textwidth]{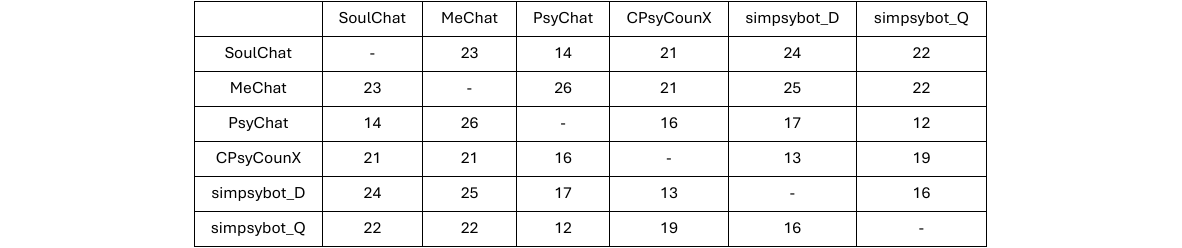}
    \caption{The times of battles between each pair of counselors.}
    \label{Fig-counselor-arena}
\end{figure*}

We first collaborate with four professional counselors to design two types of prompts for response selection, as shown in Figures~\ref{Fig-first-response-selection} and~\ref{Fig-second-response-selection}. Second, we build an online arena platform where three professional counselors select the better response from two shuffled candidate responses given a dialogue history. Each counselor annotates 97 multi-turn dialogues. Table~\ref{Tab-MRC-dialogue-statistics} and Figure~\ref{Fig-counselor-arena} present the statistics of the human-annotated dialogues in the multi-response competition task. Third, we use Qwen1.5-110B-Chat as a judge model to evaluate the response-selection accuracy of Prompt 1 and Prompt 2. Prompt 1 achieves an accuracy of 60.8\%, while Prompt 2 achieves an accuracy of 61.8\%. Furthermore, we use GPT-4o as a judge model to evaluate the response-selection accuracy of Prompt 2, achieving an accuracy of 64.4\%. Fourth, we use Prompt 2 as the response-selection guideline for Qwen1.5-110B-Chat and GPT-4o in automatic multi-response competition.

\begin{table*}[ht]
\centering
\scalebox{0.7}{
\begin{tabular}{ccccccc}
\toprule
\textbf{Category}                     & \textbf{Total} & \textbf{Client} & \textbf{Counselor} & \textbf{Total Upvotes} & \textbf{No. of Occurrence} & \textbf{Average Upvotes} \\ \hline
\textbf{\# Dialogues} & 291   & -  & - & - & - & - \\
\textbf{\# Utterances} & 7134 & 3567  & 3567 & - & - & -           \\
\textbf{Turns per dialogue}   & 12.3    & - & - & - & - & -   \\ \hline
\texttt{SoulChat}   & - & - & - & 471 & 104 & 4.53  \\
\texttt{MeChat}   & - & - & - & 705 & 117 & 6.03  \\
\texttt{PsyChat}   & - & - & - & 541 & 85 & 6.36  \\
\texttt{CPsyCounX}   & - & - & - & 418 & 90 & 4.64 \\
\texttt{simpsybot\_D}   & - & - & - & 808 & 95 & 8.51 \\
\texttt{simpsybot\_Q}   & - & - & - & 724 & 91 & 7.95  \\

\bottomrule      
\end{tabular}
}
\caption{Data statistics of the dialogue dataset in the multi-response competition task labeled by professional counselors.}
\label{Tab-MRC-dialogue-statistics}
\end{table*}

\begin{figure*}[ht]
    \centering
    \includegraphics[width=\textwidth]{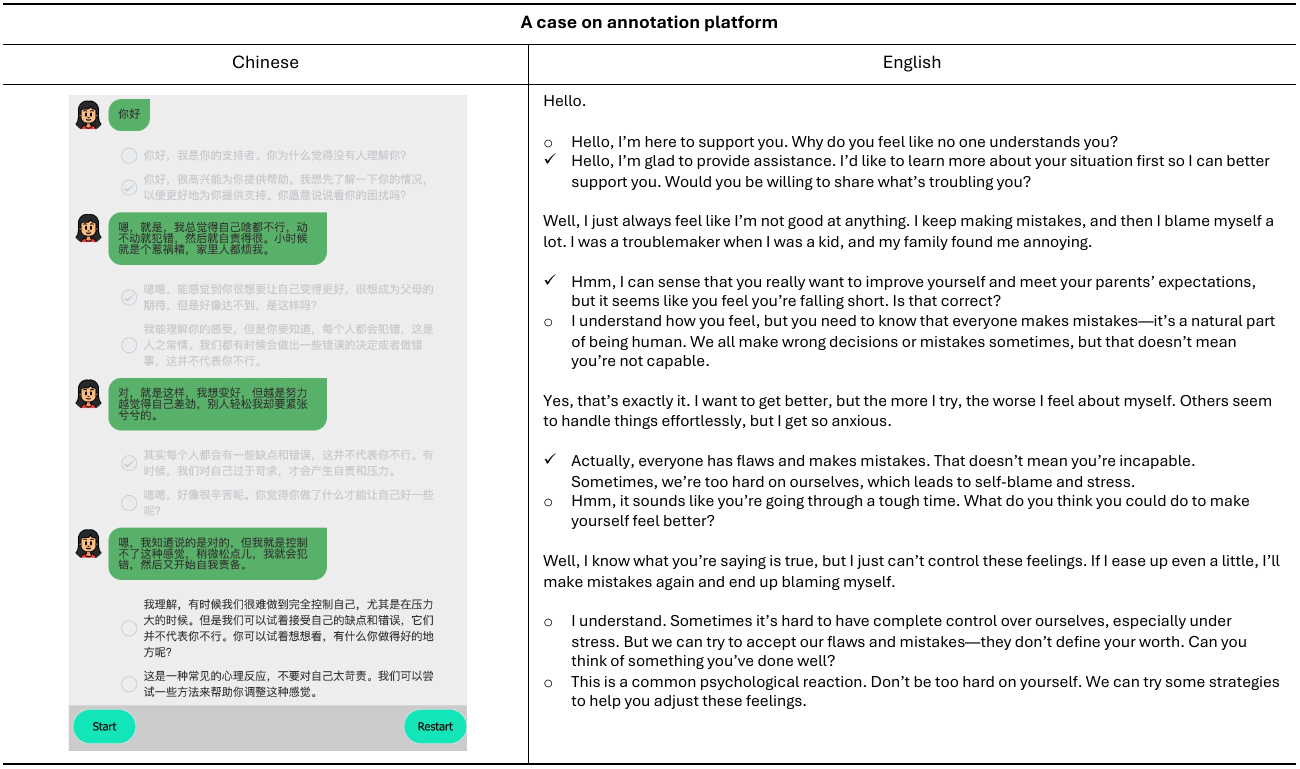}
    \caption{Snapshot of annotation platform.}
    \label{Fig-snapshot-of-ap}
\end{figure*}

\end{document}